\documentclass[10pt,twocolumn,letterpaper]{article}

\usepackage{cvpr}
\usepackage{times}
\usepackage{epsfig}
\usepackage{graphicx}
\usepackage{amsmath}
\usepackage{caption}
\usepackage{amssymb}
\usepackage{acronym} 
\usepackage{pgfplots}
\usepackage[symbol]{footmisc}

\newcommand{\modelname}{VOCA\xspace}
\newcommand{\databasename}{VOCASET\xspace}

\newcommand\blfootnote[1]{%
	\begingroup
	\renewcommand\thefootnote{}\footnote{#1}%
	\addtocounter{footnote}{-1}%
	\endgroup
}

\newcommand{\qheading}[1]{\textbf{#1}}
\renewcommand{\thefootnote}{\fnsymbol{footnote}}

\acrodef{asr}[ASR]{Automatic Speech Recognition}
\acrodef{rnn}[RNN]{Recurrent Neural Network}
\acrodef{amt}[AMT]{Amazon Mechanical Turk}

\acrodef{hmm}[HMM]{Hidden Markov Model}
\acrodef{lpc}[LPC]{Linear Predictive Coding}
\acrodef{mfcc}[MFCC]{Mel-frequency Cepstral Coefficients}
\acrodef{pca}[PCA]{Principal Component Analysis}
\acrodef{lstm}[LSTM]{Long Short-term Memory}
\acrodef{aam}[AAM]{active appearance model}
\acrodef{hit}[HIT]{human intelligent task}

\definecolor{chartblue}{RGB}{30, 120, 178}
\definecolor{chartorange}{RGB}{253, 127, 40}

\pgfplotsset{
  tick label style = {font=\sffamily},
  every axis label = {font=\sffamily},
  legend style = {font=\sffamily},
  label style = {font=\sffamily}
}
\tikzset{every picture/.style={/utils/exec={\sffamily\small}}}

\usepackage[pagebackref=true,breaklinks=true,letterpaper=true,colorlinks,bookmarks=false]{hyperref}

\cvprfinalcopy 

\def\cvprPaperID{4443} 

\begin{document}

\title{Capture, Learning, and Synthesis of 3D Speaking Styles}

\author{Daniel Cudeiro$^{*}$\textsuperscript{\textdagger} \qquad Timo Bolkart$^*$ \qquad Cassidy Laidlaw \\  Anurag Ranjan \qquad Michael J. Black\\ 
	Max Planck Institute for Intelligent Systems, T\"ubingen, Germany\\
	{\tt\small \{tbolkart, claidlaw, aranjan, black\}@tuebingen.mpg.de}
}

\twocolumn[{%
	\renewcommand\twocolumn[1][]{#1}%
	\maketitle
	\begin{center}
		\centering
		\centerline{
			\includegraphics[width=0.15\linewidth]{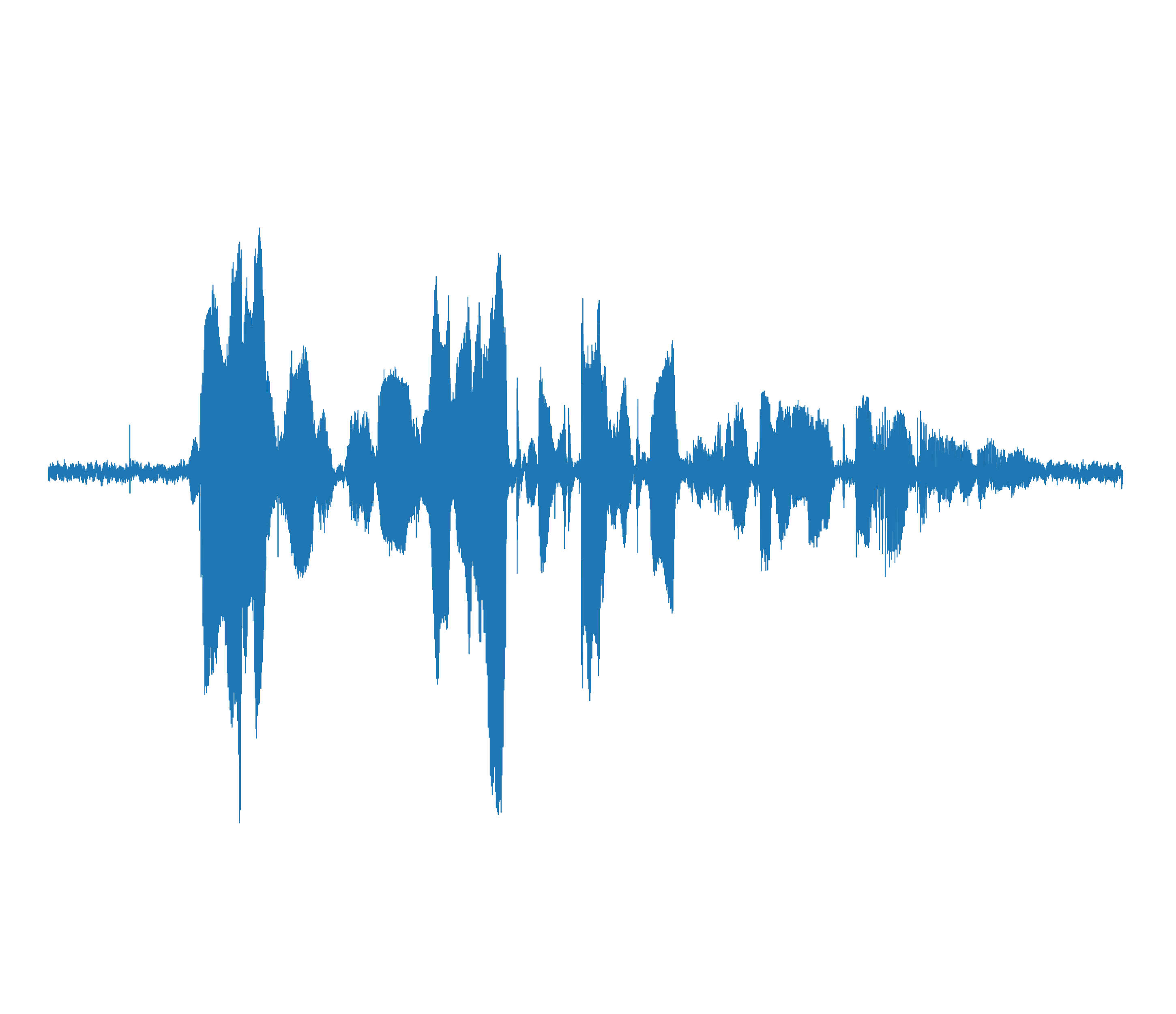}
			\includegraphics[width=0.12\linewidth]{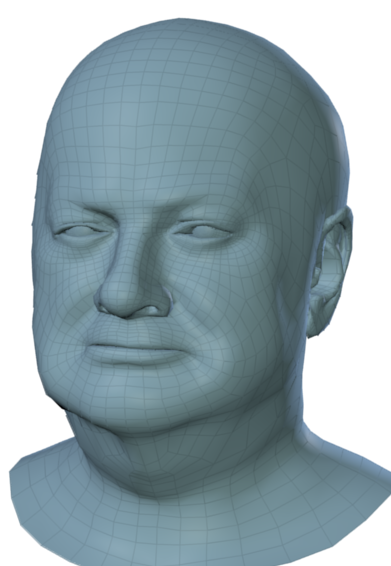}
			\hspace{0.05\linewidth}
			\includegraphics[width=0.12\linewidth]{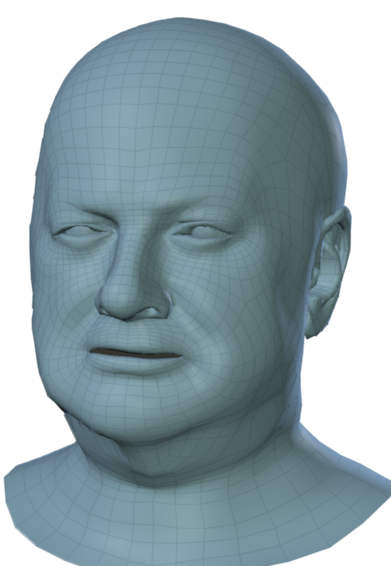}
			\includegraphics[width=0.12\linewidth]{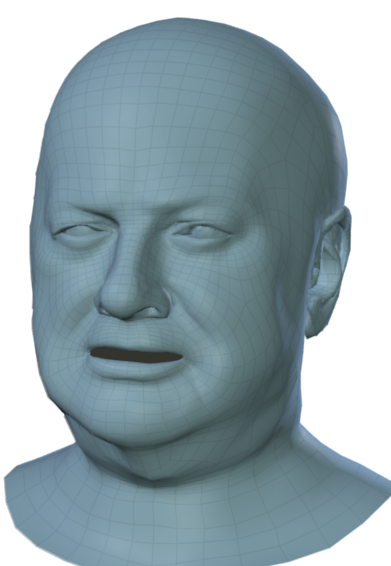}
			\includegraphics[width=0.12\linewidth]{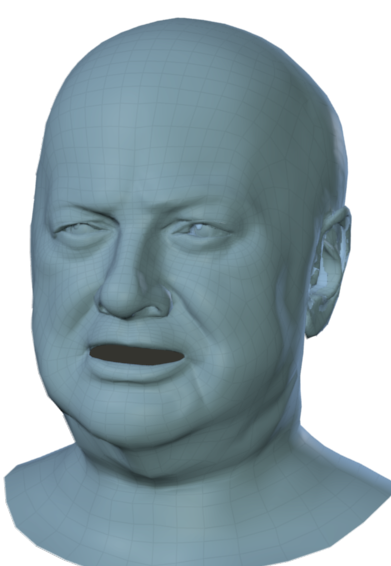}
			\includegraphics[width=0.12\linewidth]{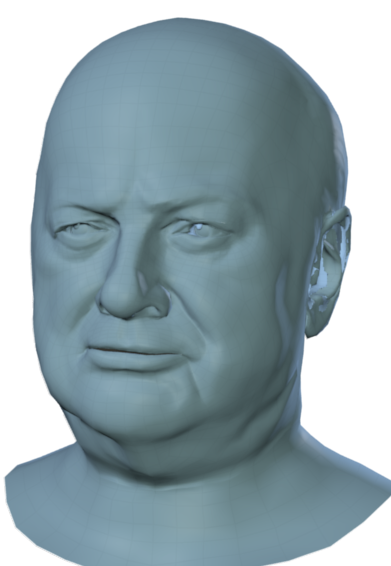}
			\includegraphics[width=0.12\linewidth]{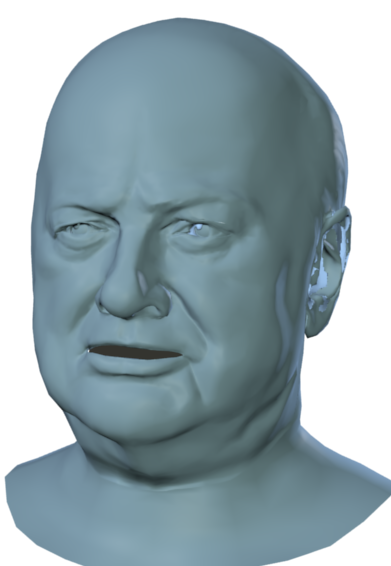}
		}
		\centerline{
			\includegraphics[width=0.15\linewidth]{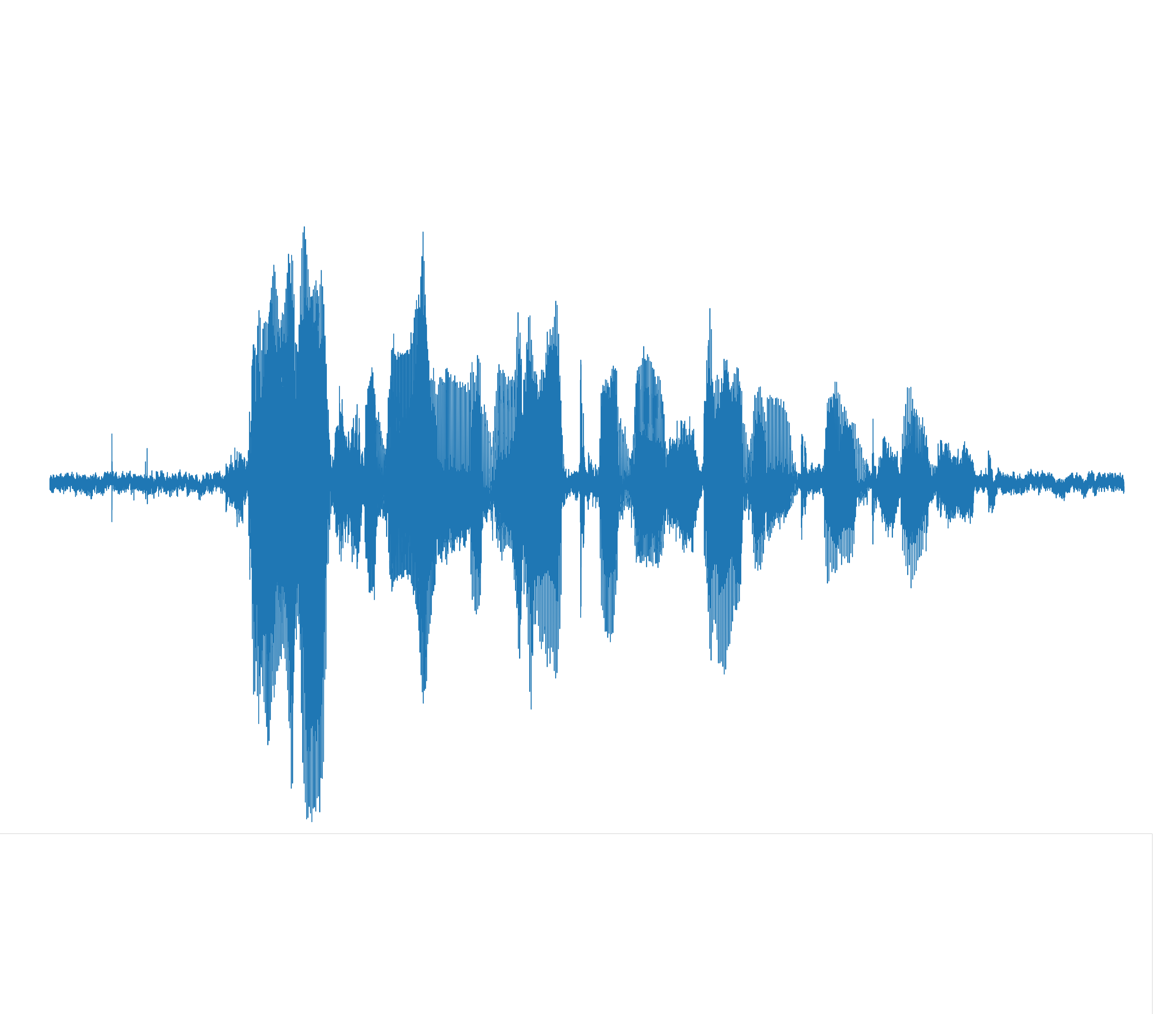}
			\includegraphics[width=0.12\linewidth]{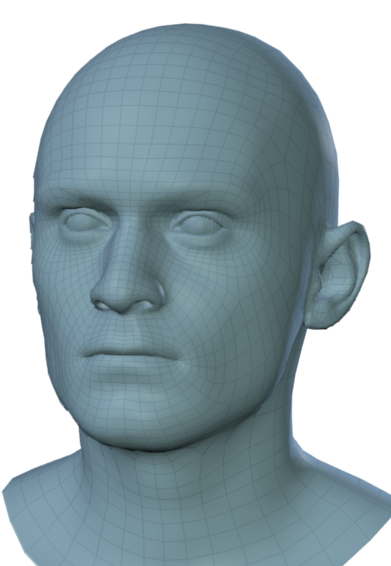}
			\hspace{0.05\linewidth}
			\includegraphics[width=0.12\linewidth]{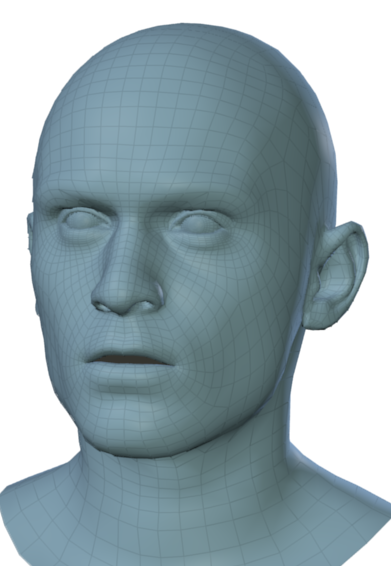}
			\includegraphics[width=0.12\linewidth]{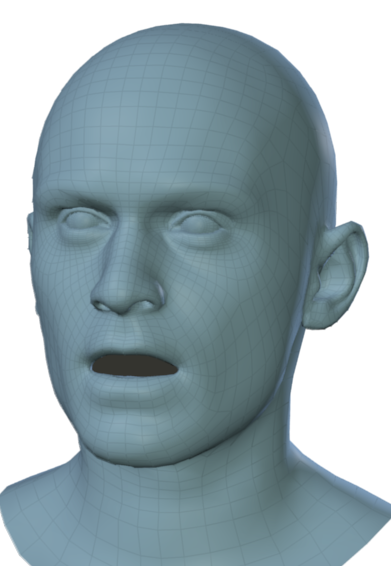}
			\includegraphics[width=0.12\linewidth]{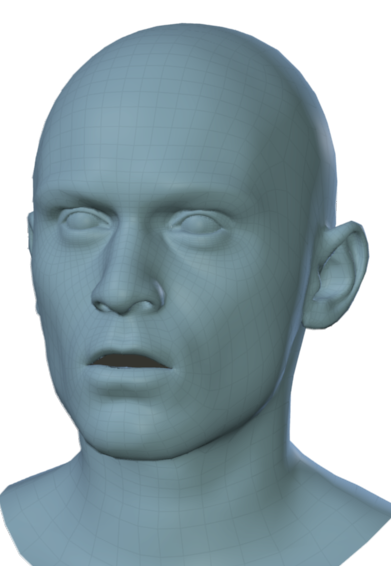}
			\includegraphics[width=0.12\linewidth]{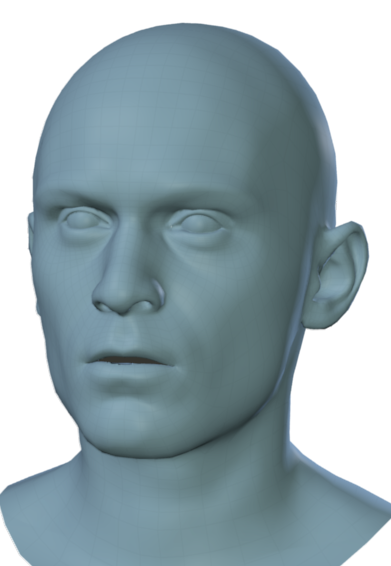}
			\includegraphics[width=0.12\linewidth]{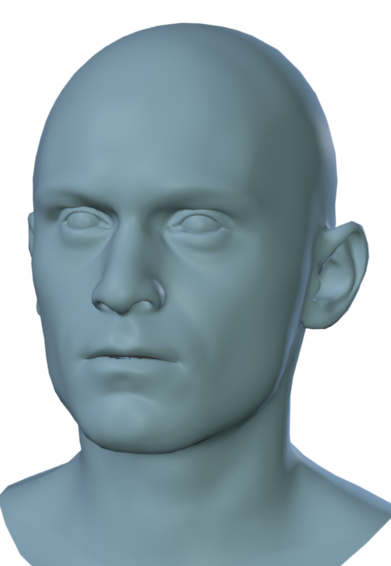}
		}
		\centerline{
			\textbf{Input}: speech signal and 3D template
			\hspace{0.18\linewidth}
			\textbf{Output}: 3D character animation
			\hspace{0.18\linewidth}
		}
		\captionof{figure}{Given an arbitrary speech signal
			and a static 3D face mesh as input (left), our model,
			\modelname outputs a realistic 3D
			character animation (right). Top: Winston
			Churchill. Bottom: Actor from Karras et
			al.~\cite{karras2017}. See {\bf supplementary video}.}
		\label{fig:teaser}
	\end{center}%
}]

\begin{abstract}

Audio-driven 3D facial animation has been widely explored, but achieving realistic, human-like performance is still unsolved.
This is due to the lack of available 3D datasets, models, and standard evaluation metrics.
To address this, we introduce a unique 4D face dataset with about 29 minutes of 4D scans captured at 60 fps and synchronized audio from 12 speakers.
We then train a neural network on our dataset that factors identity from facial motion.
The learned model, VOCA (Voice Operated Character Animation) takes any speech signal as input---even speech in languages other than English---and realistically animates a wide range of adult faces.
Conditioning on subject labels during training allows the model to learn a variety of realistic speaking styles. 
\modelname also provides animator controls to alter speaking style, identity-dependent facial shape, and pose (i.e. head, jaw, and eyeball rotations) during animation.
To our knowledge, \modelname is the only realistic 3D facial animation model that is readily applicable to unseen subjects without retargeting.
This makes \modelname suitable for tasks like in-game video, virtual reality avatars, or any scenario in which the speaker, speech, or language is not known in advance. 
We make the dataset and model available for research purposes at \url{http://voca.is.tue.mpg.de}.
\blfootnote{$^*$ Equal contribution}
\blfootnote{\textsuperscript{\textdagger} Deceased, December 5, 2018}

\end{abstract}


\newcommand{\shapecoeff}{\vec{\beta}}
\newcommand{\shapedim}{\vec{ \left| \beta \right|} }
\newcommand{\shapespace}{\mathcal{S}}
\newcommand{\posecoeff}{\vec{\theta}}
\newcommand{\posedim}{\vec{ \left| \theta \right|} }
\newcommand{\posespace}{\mathcal{P}}
\newcommand{\expcoeff}{\vec{\psi}}
\newcommand{\expdim}{\vec{ \left| \psi \right|}}
\newcommand{\expspace}{\mathcal{E}}
\newcommand{\template}{\textbf{T}}
\newcommand{\joints}{\textbf{J}}
\newcommand{\jointregressor}{\mathcal{J}}
\newcommand{\blendweights}{\mathcal{W}}
\newcommand{\blendweightsdim}{\left| \mathcal{W} \right|}
\section{Introduction}

Teaching computers to see and understand faces is critical for them to understand human behavior.
There is an extensive literature on estimating 3D face shape, facial expressions, and facial motion from images and videos.
Less attention has been paid to estimating 3D properties of faces from sound; however, many facial motions are caused directly by the production of speech.
Understanding the correlation between speech and facial motion thus provides additional valuable information for analyzing humans, particularly if visual data are noisy, missing, or ambiguous.
The relation between speech and facial motion has previously been used to separate audio-visual speech~\cite{Ephrat:Siggraph:2018} and for audio-video driven facial animation~~\cite{Liu2015}.
Missing to date is a general and robust method that relates the speech of {\em any} person in {\em any} language to the 3D facial motion of {\em any} face shape.
Here we present {\em VOCA (Voice Operated Character Animation)}, that takes a step towards this goal.

While speech-driven 3D facial animation has been widely studied, speaker-independent modeling remains a challenging, unsolved task for several reasons.
First, speech signals and facial motion are strongly correlated but lie in two very different spaces; thus, non-linear regression functions are needed to relate the two.
One can exploit deep neural networks to address this problem.
However, this means that significant amounts of training data are needed.
Second, there exists a many-to-many mapping between phonemes and facial motion.
This poses an even greater challenge when training across people and styles.
Third, because we are especially sensitive to faces, particularly realistic faces, the animation must be realistic to avoid falling into the Uncanny Valley~\cite{Mori1970}.
Fourth, there is very limited training data relating speech to the 3D face shape of multiple speakers.
Finally, while previous work has shown that models can be trained to create speaker-specific animations~\cite{Cao2005, karras2017}, there are no generic methods that are speaker independent and that capture a variety of speaking styles.

\qheading{\databasename:} To address this, we collected a new dataset of 4D face scans together with speech. 
The dataset has 12 subjects and 480 sequences of about 3-4 seconds each with sentences chosen from an array of standard protocols that maximize phonetic diversity.
The 4D scans are captured at 60fps and we align a common face template mesh to all the scans, bringing them into correspondence.
This dataset, called {\em \databasename}, is unlike any existing public datasets.
It allows training and testing of speech-to-animation models that can generalize to new data.

\qheading{\modelname:} Given such data, we train a deep neural network model, called {\em \modelname} (Figure~\ref{fig:network-architecture}), that generalizes to new speakers (see Figure~\ref{fig:teaser}).
Recent work using deep networks has shown impressive results for the problem of regressing {\em speaker-dependent} facial animation from speech \cite{karras2017}.
Their work, however, captures the idiosyncrasies of an individual, making it inappropriate for generalization across characters.
While deep learning is advancing the field quickly, even the best recent methods rely on some manual processes or focus only on the mouth \cite{taylor2017deep}, making them inappropriate for truly automatic full facial animation.

The key problem with prior work is that facial motion and facial identity are confounded.
Our key insight is to factor identity from facial motions and then learn a model relating speech to only the motions.
Conditioning on subject labels during training allows us to combine data from many subjects in the training process, which enables the model both to generalize to new subjects not seen during training and to synthesize different speaker styles.
Integrating DeepSpeech~\cite{hannun2014deep} for audio feature extraction makes \modelname robust w.r.t. different audio sources and noise.
Building on top of the expressive FLAME head model~\cite{flame2017} allows us i) to model motions of the full face (i.e. including the neck), ii) to animate a wide range of adult faces, as FLAME can be used to reconstruct subject-specific templates from a scan or image, and iii) to edit identity-dependent shape and head pose during animation.
\modelname and \databasename are available for research purposes~\cite{VOCA_project}.

\section{Related work} \label{related}

Facial animation has received significant attention in the literature. Related work in this area can be grouped into three categories: speech-based, text-based, and video- or performance-based.

\qheading{Speech-driven facial animation:} Due to the abundance of images and videos, many methods that attempt to realistically animate faces use monocular video~\cite{Brand1999,Bregler1997,Chen2018,Ezzat2002,obama2017,Wang2011,XieLiu2007}.
Bregler et al.~\cite{Bregler1997} transcribe speech with a \ac{hmm} into phonetic labels and animate the mouth region in videos with an exemplar-based video warping.
Brand~\cite{Brand1999} uses a mix of \ac{lpc} and RASTA-PLP~\cite{Hermansky1994} audio features and an \ac{hmm} to output a sequence of facial motion vectors.
Ezzat et al.~\cite{Ezzat2002} perform \ac{pca} on all images and use an example-based mapping between phonemes and trajectories of mouth shape and mouth texture parameters in the \ac{pca} space.
Xie and Liu~\cite{XieLiu2007} model facial animation with a dynamic Bayesian network-based model.
Wang et al.~\cite{Wang2011} use an \ac{hmm} to learn a mapping between \ac{mfcc} and \ac{pca} model parameters.
Zhang et al.~\cite{Zhang2013} combine the \ac{hmm}-based method of \cite{Wang2011} trained on audio and visual data of one actor with a deep neural network based encoder trained from hundreds of hours of speaker independent speech data to compute an embedding of the MFCC audio features.
Shimba et al.~\cite{shimba2015talking} use a deep \ac{lstm} network to regress \ac{aam} parameters from \ac{mfcc} features.
Chen et al.~\cite{Chen2018} correlate audio and image motion to synthesize lip motion of arbitrary identities.

Suwajanakorn et al.~\cite{obama2017} use an \ac{rnn} for synthesizing photorealistic mouth texture animations using audio from 1.9 million frames from Obama's weekly addresses. However, their method does not generalize to unseen faces or viewpoints.
In contrast to this, \modelname is trained across subjects sharing a common topology, which makes it possible to animate new faces from previously unseen viewpoints.
Pham et al.~\cite{Pham2017} regress global transformation and blendshape coefficients~\cite{Cao2014_FaceWarehouse} from MFCC audio features using an \ac{lstm} network. While their model is trained across subjects---similar to \modelname---they rely on model parameters regressed from 2D videos rather than using 3D scans, which limits their quality.

A few methods use multi-view motion capture data~\cite{Busso2007,Cao2005} or high-resolution 3D scans~\cite{karras2017}. Busso et al.~\cite{Busso2007} synthesize rigid head motion in expressive speech sequences. Cao et al.~\cite{Cao2005} segment the audio into phonemes and use an example-based graph method to select a matching mouth animation.
Karras et al.~\cite{karras2017} propose a convolutional model for mapping \ac{lpc} audio features to 3D vertex displacements.
However, their model is subject specific, and animating a new face would require 3D capture and processing of thousands of frames of subject data.
Our model, \modelname factors identity from facial motion and is trained across subjects, which allows animation of a wide range of adult faces.

Several works also aim at animating artist designed character rigs~\cite{Ding2015,Edwards2016_JALI,Hong2002,kakumanu2001speech,Salvi2009,taylor2016audio,taylor2017deep,taylor2012dynamic,Zhou2018}.
Taylor et al.~\cite{taylor2017deep} propose a deep-learning based speech-driven facial animation model using a sliding window approach on transcribed phoneme sequences that outperforms previous LSTM based methods~\cite{fan2015photo, fan2016deep}.
While these models are similar to \modelname in that they animate a generic face from audio, our focus is animating a realistic face mesh, for which we train our model on high-resolution face-scans.

\qheading{Text-driven facial animation:}
Some methods aim to animate faces directly from text. Sako et al.~\cite{Sako2000} use a hidden Markov model to animate lips in images from text. Anderson et al.~\cite{Anderson2013} use an extended hidden Markov text-to-speech model to drive a subject-specific active appearance model (AAM). In a follow-up, they extend this approach to animate the face of an actress in 3D. While our focus is not to animate faces from text, this is possible by animating our model with the output of a text-to-speech (TTS) system (e.g.~\cite{WaveNet2016}), similar to Karras et al.~\cite{karras2017}.

\qheading{Performance-based facial animation:} Most methods to animate digital avatars are based on visual data. Alexander et al.~\cite{Alexander2009}, Wu et al.~\cite{Wu2016}, and Laine et al.~\cite{Laine2017} build subject-specific face-rigs from high-resolution face scans and animate these rigs with video-based animation systems.

Several methods build personalized face-rigs using generic face models from monocular videos to transfer and reenact facial performance between videos. Tensor-based multilinear face models~\cite{Bolkart2015,Cao2015,Cao2014,Dale2011,Vlasic2005,Yang2012} and linear models~\cite{Thies2016_Face2Face} are widely used to build personalized face-rigs. Cao et al.~\cite{Cao2015,Cao2014} use a regression-based face tracker to animate the face-rig and digital avatars, while Thies et al.~\cite{Thies2016_Face2Face} use a landmark-based face tracker and deformation transfer~\cite{SumnerPopovic2004} to reenact monocular videos.

Other methods that animate virtual avatars rely on {RGB-D} videos or 4D sequences to track and retarget facial performance.
Li et al.~\cite{Li2010} and Weise et al.~\cite{Weise2011} capture example-based rigs in an offline calibration procedure to build personalized face-rigs, Bouaziz et al.~\cite{Bouaziz2013} use a generic identity model.
Liu et al.~\cite{Liu2015} combine audio and video to robustly animate a generic face model from RGB-D video.
Li et al.~\cite{flame2017} capture facial performance with a high-resolution scanner and animate static face meshes using an articulated generic head model.
In contrast to these methods, our approach solely relies on audio to animate digital avatars.

\qheading{3D face datasets:} Several 3D face datasets have been released that focus on the analysis of static 3D facial shape and expression  (e.g.~\cite{Cao2014_FaceWarehouse,Bosphorus2008,BU-3DFE2006}) or dynamic facial expressions (e.g.~\cite{Alashkar2014,Chang2005,D3DFACS2011,CoMA2018,BU-4DFE2008,BP4D-Spontaneous2014,MMSE2016}).
Most of these datasets focus on emotional expressions and only a few datasets  capture facial dynamics caused by speech.
The recently published 4DFAB dataset~\cite{4DFAB2018} contains 4D captures of 180 subjects, but with only nine word utterances per subject and lower mesh quality than VOCASET.

The B3D(AC)\^{}2 dataset~\cite{B3D(AC)2010} contains a large set of audio-4D scan pairs of 40 spoken English sentences. In contrast, \databasename contains 255 unique sentences in total. To enable training on both a large number of sentences and subjects, some sentences are shared across subjects and some sentences are spoken by only one subject.
The visible artifacts present in the raw B3D(AC)\^{}2 scans (i.e.~holes and capture noise) mean that subtle facial motions may be lost; also, the registered template only covers the face, ignoring speech-related motions in the neck region. \databasename, in comparison, provides higher-quality 3D scans as well as alignments of the entire head, including the neck.

\section{Preliminaries}

Our goal for \modelname is to generalize well to arbitrary subjects not seen during training. Generalization across subjects involves both (i) generalization across different speakers in terms of the audio (variations in accent, speed, audio source, noise, environment, etc.) and (ii) generalization across different facial shapes and motion. 

\qheading{DeepSpeech:} To gain robustness to different audio sources, regardless of noise, recording artifacts, or language, we integrate DeepSpeech~\cite{hannun2014deep} into our model. DeepSpeech~\cite{hannun2014deep} is an end-to-end deep learning model for \ac{asr}. 
DeepSpeech uses a simple architecture consisting of five layers of hidden units, of which the first three layers are non-recurrent fully connected layers with ReLU activations. The fourth layer is a bi-directional \ac{rnn}, and the final layer is a fully connected layer with ReLU activation. The final layer of the network is fed to a softmax function whose output is a probability distribution over characters. The TensorFlow implementation provided by Mozilla~\cite{mozillaDeepSpeech} slightly differs from the original paper in two ways: (i) the \ac{rnn} units are replaced by \ac{lstm} cells and (ii) 26 \ac{mfcc} audio features are used instead of directly performing inference on the spectrogram. Please see \cite{mozillaDeepSpeech} for more details.

\qheading{FLAME:} Facial shape and head motion vary greatly across subjects. Furthermore, different people have different speaking styles. The large variability in facial shape, motion, and speaking style motivates using a common learning space. We address this problem by incorporating FLAME, a publicly available statistical head model, as part of our animation pipeline. FLAME uses linear transformations to describe identity and expression dependent shape variations, and standard linear blend skinning (LBS) to model neck, jaw, and eyeball rotations. Given a template $\template \in \mathbb{R}^{3N}$ in the ``zero pose'', identity, pose, and expression blendshapes are modeled as vertex offsets from $\template$. For more details we refer the reader to~\cite{flame2017}.

\section{VOCA} \label{model}

\begin{figure*}
	\centering
	\includegraphics[width=0.8\linewidth]{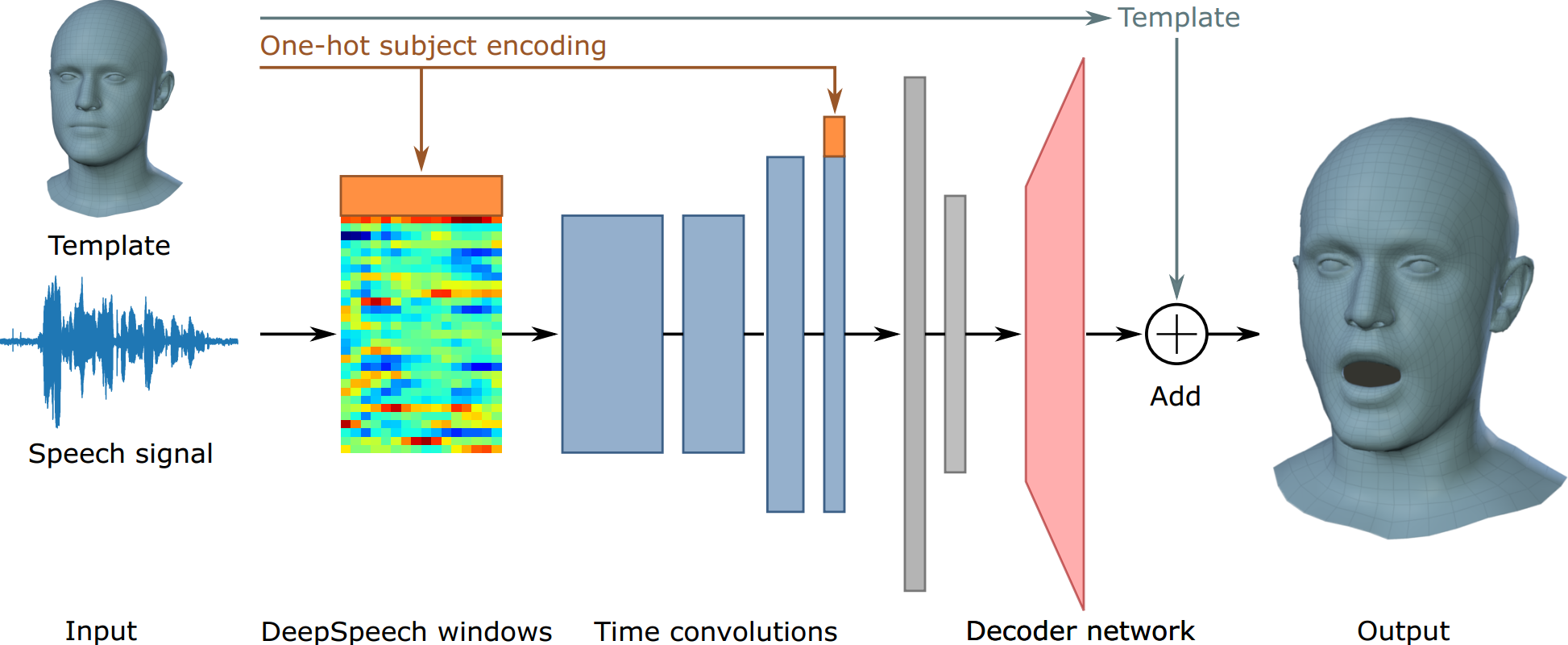}
	\caption{\modelname network architecture.}
	\label{fig:network-architecture}
\end{figure*} 

\begin{table}[t]
	\footnotesize
	\begin{tabular}{lcccc}
		\bf Type & \bf Kernel  & \bf Stride  & \bf Output  & \bf Activation  \\
		\hline
		DeepSpeech  & - & - & 16x1x29 & - \\
		\hline
		Identity concat & - & - & 16x1x37 & - \\						
		Convolution & 3x1 & 2x1 & 8x1x32 & ReLU \\
		Convolution & 3x1 & 2x1 & 4x1x32 & ReLU \\
		Convolution & 3x1 & 2x1 & 2x1x64 & ReLU \\
		Convolution & 3x1 & 2x1 & 1x1x64 & ReLU \\
		\hline
		Identity concat & - & - & 72 & - \\
		Fully connected & - & - & 128 & tanh \\
		Fully connected & - & - & 50 & linear \\
		\hline
		Fully connected	& - & - & 5023x3 & linear \\
	\end{tabular}
	\caption{Model architecture.}
	\label{tab:architecture}
\end{table}

This section describes the model architecture and provides details on how the input audio is processed.

\qheading{Overview:} \modelname receives as input a subject-specific template $\template$ and the raw audio signal, from which we extract features using DeepSpeech~\cite{hannun2014deep}. The desired output is the target 3D mesh. \modelname acts as an encoder-decoder network (see Figure~\ref{fig:network-architecture} and Table~\ref{tab:architecture}) where the encoder learns to transform audio features to a low-dimensional embedding and the decoder maps this embedding into a high-dimensional space of 3D vertex displacements

\qheading{Speech feature extraction:} Given an input audio clip of length $T$ seconds, we use DeepSpeech to extract speech features. The outputs are unnormalized log probabilities of characters for frames of length 0.02 s (50 frames per second); thus, it is an array of size $50 T \times D$, where $D$ is the number of characters in the alphabet plus one for a blank label. We resample the output to 60 fps using linear interpolation. In order to incorporate temporal information, we convert the audio frames to overlapping windows of size $W \times D$, where $W$ is the window size. The output is a three-dimensional array of dimensions $60 T \times W \times D$.

\qheading{Encoder:} The encoder is composed of four convolutional layers and two fully connected layers. The speech features and the final convolutional layer are conditioned on the subject labels to learn subject-specific styles when trained across multiple subjects. For eight training subjects, each subject $j$ is encoded as an one-hot-vector $I_j = \left(\delta_{ij}\right)_{1 \leq i \leq 8}$. This vector is concatenated to each $D$-dimensional speech feature vector (i.e. resulting in windows of dimension $W \times (D+8)$), and concatenated to the output of the final convolution layer. 

To learn temporal features and reduce the dimensionality of the input, each convolutional layer uses a kernel of dimension $3 \times 1$ and stride $2 \times 1$. As the features extracted using DeepSpeech do not have any spatial correlation, we reshape the input window to have dimensions $W \times 1 \times (D+8)$ and perform 1D convolutions over the temporal dimension. To avoid overfitting, we keep the number of parameters small and only learn 32 filters for the first two, and 64 filters for the last two convolutional layers.
 
The concatenation of the final convolutional layer with the subject encoding is followed by two fully connected layers. The first has 128 units and a hyperbolic tangent activation function; the second is a linear layer with 50 units.

\qheading{Decoder:} The decoder of \modelname is a fully connected layer with linear activation function, outputting the $5023 \times 3$ dimensional array of vertex displacements from $\template$. The weights of the layer are initialized by $50$ PCA components computed over the vertex displacements of the training data; the bias is initialized with zeros. 

\qheading{Animation control:} During inference, changing the eight-dimensional one-hot-vector alters the output speaking style. The output of \modelname is an expressed 3D face in ``zero pose'' with the same mesh topology as the FLAME face model \cite{flame2017}. \modelname's compatibility with FLAME allows alteration of the identity-dependent facial shape by adding weighted shape blendshapes from FLAME. The face expression and pose (i.e. head, jaw, and eyeball rotations) can also be changed using the blendweights, joints, and pose blendshapes provided by FLAME.

\section{Model training} \label{training}

In this section we describe training relevant details. 

\qheading{Training set-up:} We start from a large dataset of audio-4D scan pairs, denoted as $\{(\textbf{x}_{i}, \textbf{y}_{i})\}_{i = 1}^{F}$. Here $\textbf{x}_{i} \in  \mathbb{R}^{W \times D}$ is the input audio window centered at the $i$th video frame, $\textbf{y}_{i} \in \mathbb{R}^{N\times3}$. Further, let $\textbf{f}_i \in \mathbb{R}^{N\times3}$ denote the output of \modelname for $\textbf{x}_{i}$. 

For training, we split the captured data into a training set (eight subjects), a validation set (two subjects), and a test set (two subjects). The training set consists of all 40 sentences of the eight subjects, i.e. in total 320 sentences. 
For validation and test data, we only select the 20 unique sentences that are not shared with any other subject, i.e. 40 sentences for validation and testing, respectively. Note that our training, validation, and test sets for all experiments are fully disjoint, i.e. no overlap of subjects or sentences exists. 

\qheading{Loss function:} Our training loss function consists of two terms, a position term and a velocity term. The position term $E_p = \Vert \textbf{y}_{i} - \textbf{f}_i \Vert_{F}^2$ computes the distance between the predicted outputs and the training vertices. This position term encourages the model to match the ground truth performance. The velocity term $E_v = \Vert (\textbf{y}_{i} - \textbf{y}_{i-1}) - (\textbf{f}_i - \textbf{f}_{i-1})\Vert_{F}^2$ uses backward finite differences. It computes the distance between the differences of consecutive frames between predicted outputs and training vertices. This velocity term induces temporal stability.

\qheading{Training parameters:} We perform hyperparameter tuning on the held-out validation set. We train \modelname for 50 epochs with a constant learning rate of $1e-4$. The weights for the position and velocity terms are $1.0$ and $10.0$, respectively. During training, we use batch normalization with a batch size of 64. We use a window size of $W=16$ with $D=29$ speech features.

\qheading{Implementation details: } \modelname is implemented in Python using TensorFlow \cite{Abadi2016}, and trained using Adam \cite{kingma2014adam}. Training one epoch takes about ten minutes on a single NVIDIA Tesla K20. We use a pre-trained DeepSpeech model~\cite{mozillaDeepSpeech} which is kept fixed during training.

\section{VOCASET} \label{data}

This section introduces \databasename and describes the capture setup and data processing.

\qheading{\databasename:} Our dataset contains a collection of audio-4D scan pairs captured from 6 female and 6 male subjects. For each subject, we collect 40 sequences of a sentence spoken in English, each of length three to five seconds. The sentences were taken from an array of standard protocols and were selected to maximize phonetic diversity using the method described in \cite{fisher1986}. In particular, each subject spoke 27 sentences from the TIMIT corpus \cite{timit1993}, three pangrams used by \cite{karras2017}, and 10 questions from the Stanford Question Answering Dataset (SQuAD) \cite{squad2016}. The recorded sequences are distributed such that five sentences are shared across all subjects, 15 sentences are spoken by three to five subjects (50 unique sentences), and 20 sentences are spoken only by one or two subjects (200 unique sentences). We make \databasename available to the research community. 

\begin{figure}[t]
	\centerline{
		\includegraphics[width=0.24\columnwidth]{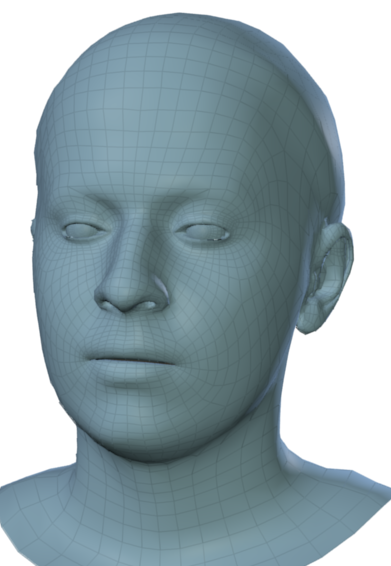} 
		\includegraphics[width=0.24\columnwidth]{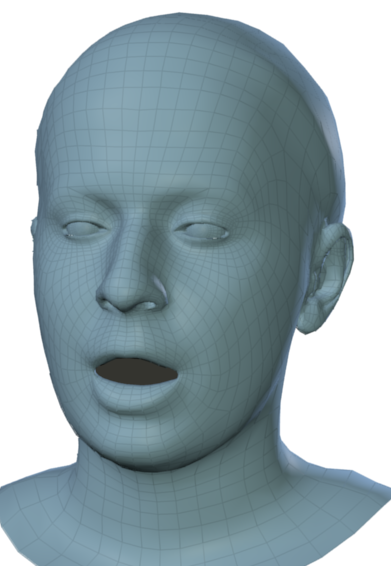} 
		\includegraphics[width=0.24\columnwidth]{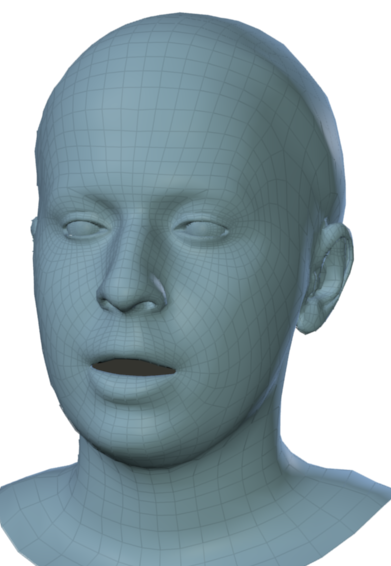} 
		\includegraphics[width=0.24\columnwidth]{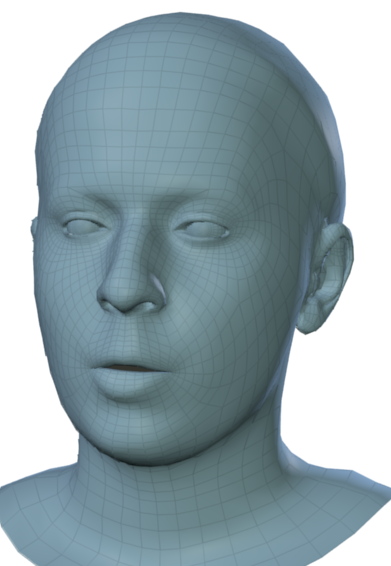} 
	}
	\centerline{	
		\includegraphics[width=0.24\columnwidth]{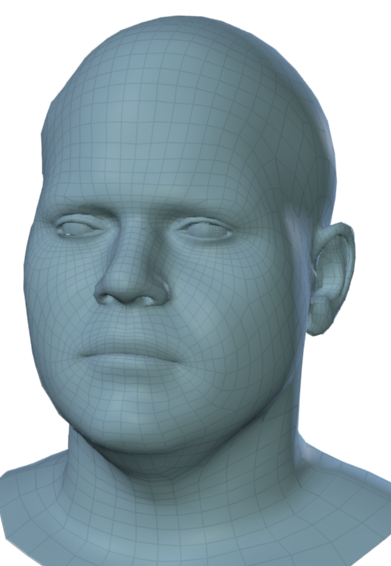} 
		\includegraphics[width=0.24\columnwidth]{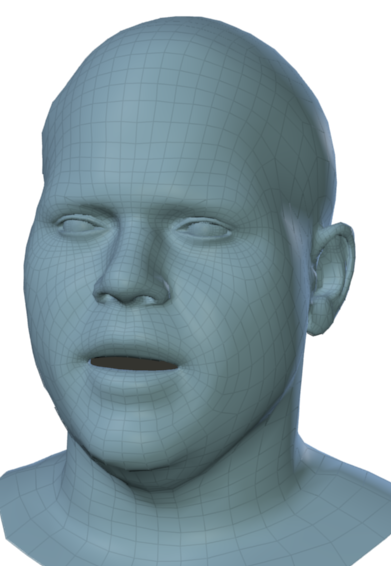} 
		\includegraphics[width=0.24\columnwidth]{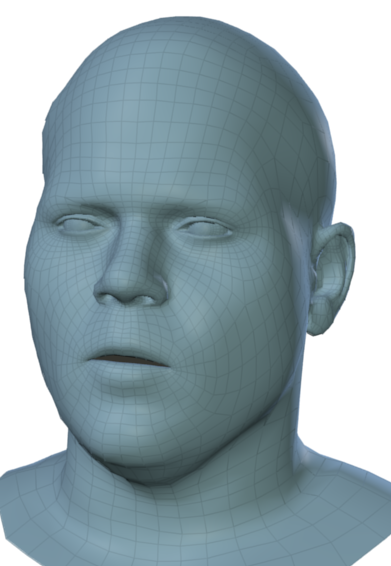} 								
		\includegraphics[width=0.24\columnwidth]{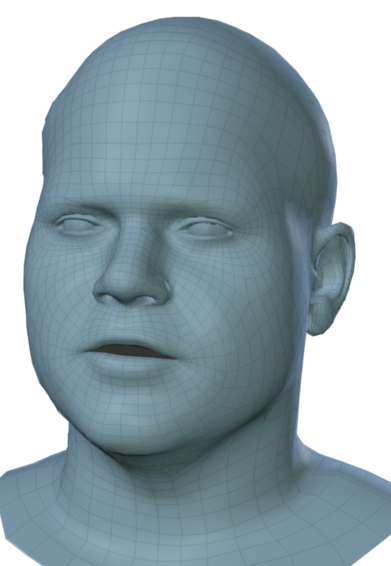} 				
	}
	\caption{Sample meshes of two \databasename subjects.}
	\label{fig:dataset}
\end{figure}

\qheading{Capture setup:} We use a multi-camera active stereo system (3dMD LLC, Atlanta) to capture high-quality 3D head scans and audio. The capture system consists of six pairs of gray-scale stereo cameras, six color cameras, five speckle pattern projectors, and six white light LED panels. The system captures 3D meshes at 60fps, each with about 120K vertices. The color images are used to generate UV texture maps for each scan. The audio, synchronized with the scanner, is captured with a sample rate of 22 kHz.

\qheading{Data processing:} The raw 3D head scans are registered with a sequential alignment method as described in \cite{flame2017} using the publicly available generic FLAME model. The image-based landmark prediction method of \cite{BulatTzimiropoulos} is used during alignment to add robustness while tracking fast facial motions. After alignment, each mesh consists of $5023$ 3D vertices. For all scans, we measure the absolute distance between each scan vertex and the closest point in the FLAME alignment surface: median (0.09mm), mean (0.13mm), and standard deviation (0.14mm). Thus, the alignments faithfully represent the raw data. 

All meshes are then unposed; i.e. effects of global rotation, translation, and head rotation around the neck are removed. After unposing, all meshes are in ``zero pose''. For each sequence, the neck boundary and the ears are automatically fixed, and the region around the eyes is smoothed using Gaussian filtering to remove capture noise. Note that no smoothing is applied to the mouth region so as to preserve subtle motions. Figure~\ref{fig:dataset} shows sample alignments of two \databasename subjects. The supplementary video shows sequences of all subjects.

\section{Experiments} 
\label{experiments}

Quantitative metrics, such as the norm on the prediction error, are not suitable for evaluating animation quality. This is because facial visemes form many-to-many mappings with speech utterances. A wide range of plausible facial motions exists for the same speech sequence, which makes quantitative evaluation intractable. Instead, we perform perceptual and qualitative evaluations. Further, our trained model is available for research purposes for direct comparisons~\cite{VOCA_project}. 

\subsection{Perceptual evaluation}

\qheading{User study:} We conduct three \ac{amt} blind user studies: i) a binary comparison between held-out test sequences and our model conditioned on all training subjects, ii) an ablation study to assess the effectiveness of the DeepSpeech features, and iii) a study to investigate the correlation between style, content, and identity. All experiments are performed on sequences and subjects fully disjoint from our training and validation set.

For binary comparisons, two videos with the same animated subject and audio clip are shown side by side. For each video pair, the participant is asked to choose the talking head that moves more naturally and in accordance with the audio. To avoid any selection bias, the order (left/right) of all methods for comparison is random for each pair.

Style comparisons are used to evaluate the learned speaking styles. Here, Turkers see three videos: one reference and two predictions. The task is to determine which of the two predictions is more similar to the reference video.

To ensure the quality of the study and remove potential outliers, we require Turkers to pass a simple qualification test before they are allowed to submit HITs. The qualification task is a simplified version of the following user study, where we show three comparisons with an obvious answer, i.e. one ground-truth sequence and one sequence with completely mismatched video and audio.

\qheading{Comparison to recorded performance:} We compare captured and processed test sequences with \modelname predictions conditioned on all eight speaker styles. 
In total, Turkers ($400$ HITs) perceived the recorded performance more natural ($83 \pm 9\%$) than the predictions ($17 \pm 9\%$), across all conditions. 
While \modelname results in realistic facial motion for the unseen subjects, it is unable to synthesize the idiosyncrasies of these subjects. 
As such, these subtle subject-specific details make the recorded sequences look more natural than the predictions.

\qheading{Speech feature ablation:} We replace the DeepSpeech features by Mel-filterbank energy features (fbank) and train a model for $50$ epochs (the same as for \modelname). Turkers ($400$ HITs) perceived the performance of \modelname with DeepSpeech more natural ($78 \pm 16\%$) than with fbank features ($22 \pm 16\%$) across all conditions. That indicates that \modelname with DeepSpeech features generalizes better to unseen audio sequences than with fbank features.

\qheading{Style comparisons:} Speech-driven facial performance varies greatly across subjects. 
However, it is difficult to separate between style (facial motion of a subject), identity (facial shape of a subject), and content (the words being said), and how these different factors influence perception.
The goal of this user study is to evaluate the speech-driven facial motion independently from identity-dependent face shape in order to understand if people can recognize the styles learned by our model.

To accomplish this, we subtract the personalized template (neutral face) from all sequences to obtain ``displacements'', then add these displacements to a single common template (randomly sampled from the FLAME shape space).
Then, for several reference sequences from the training data, we compare two \modelname predictions (on audio from the test set): one conditioned on the reference subject and one conditioned on another randomly selected subject.
We ask Turkers to select which predicted sequence is more similar in speaking style to the reference.

To explore the influence of content, we perform the experiment twice, once where the reference video and the predictions share the same sentence (spoken by different subjects) and once with different sentences.
Figure~\ref{fig:exp3_and_4} shows the results for this experiment. Results varied greatly across conditions. For some conditions, Turkers could consistently pick the sequence with the matching style (e.g. conditions 3, 4, and 5); for others, their choices were no better or worse than chance. The impact of the content was not significant for most conditions.
More research is needed to understand which factors are important for people to recognize different speaking styles, and to develop new models that more efficiently disentangle facial shape and motion.

\pgfplotsset{compat=1.11,
    /pgfplots/ybar legend/.style={
    /pgfplots/legend image code/.code={%
       \draw[##1,/tikz/.cd,yshift=-0.25em]
        (0cm,0cm) rectangle (0.6em,0.8em);},
   },
}

\begin{figure}[t]
	\centering
	\begin{tikzpicture}
	\begin{axis}[
		symbolic x coords={0, 1, 2, 3, 4, 5, 6, 7, 8, 9},
		xmin=0, xmax=9,
		xtick={1, 2, 3, 4, 5, 6, 7, 8},
		xtick style={draw=none},
		xlabel=Condition,
		yticklabel={\pgfmathprintnumber[assume math mode=true]{\tick}\%},
		ylabel=Percent of reference condition chosen,
		ymajorgrids=true,
		ytick={10, 20, 30, 40, 50, 60, 70, 80, 90},
		ymin=0, ymax=100,
		ytick style={draw=none},
		legend style={at={(0.5,1.15)},
			anchor=north, legend columns=-1, draw=lightgray},
		style={draw=lightgray},
		ybar=0pt, 
		bar width=7pt
	]
	\addplot[fill=chartblue, draw opacity=0, error bars/.cd, y dir=both, y explicit,
			error bar style={draw=black, draw opacity=1}]
		coordinates {
		(1, 25) +- (0, 13.4)
		(2, 48) +- (0, 12.6)
		(3, 87) +- (0, 8.6)
		(4, 68) +- (0, 14.5)
		(5, 80) +- (0, 10.1)
		(6, 48) +- (0, 12.6)
		(7, 55) +- (0, 12.6)
		(8, 15) +- (0, 9.0)};

	\addplot[fill=chartorange, draw opacity=0, error bars/.cd, y dir=both, y explicit,
			error bar style={draw=black, draw opacity=1}]
		coordinates {
		(1, 45) +- (0, 8.9)
		(2, 34) +- (0, 9.8)
		(3, 92) +- (0, 4.7)
		(4, 67) +- (0, 8.4)
		(5, 76) +- (0, 7.7)
		(6, 26) +- (0, 7.8)
		(7, 55) +- (0, 10.0)
		(8, 14) +- (0, 6.9)};

	\legend{Same sentences, Different sentences}
	\end{axis}
	\end{tikzpicture}
	\caption{AMT study of styles. The bars show the percentage of Turkers choosing the reference condition when the same sentence was being shown for reference and prediction, and with difference sentences.}
	\label{fig:exp3_and_4}
\end{figure}
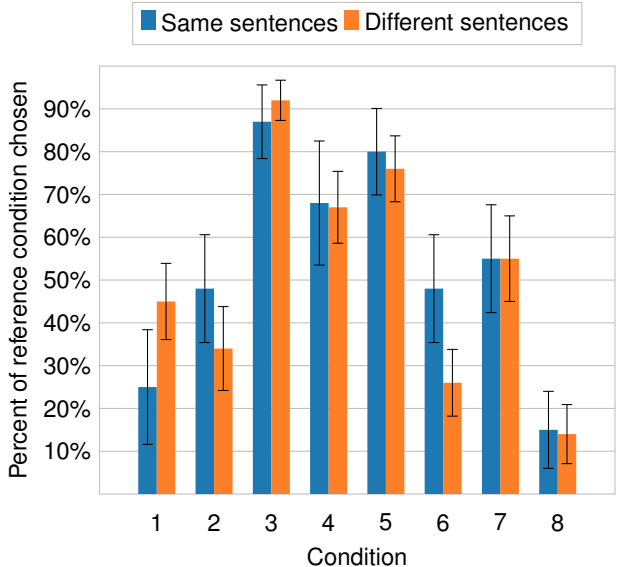

\subsection{Qualitative evaluation}

\begin{figure}[t]
	\centerline{
		\includegraphics[width=0.24\columnwidth]{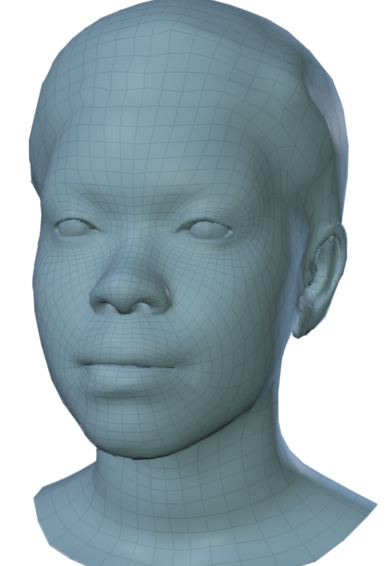} 
		\includegraphics[width=0.24\columnwidth]{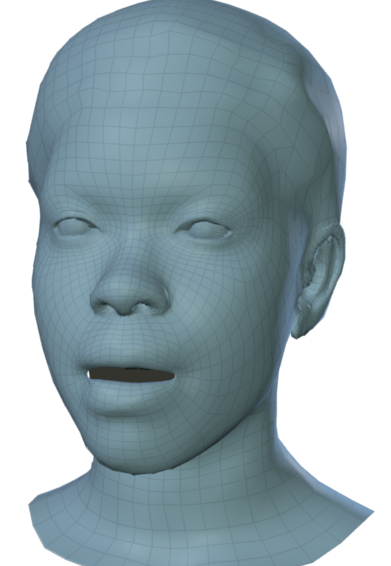} 
		\includegraphics[width=0.24\columnwidth]{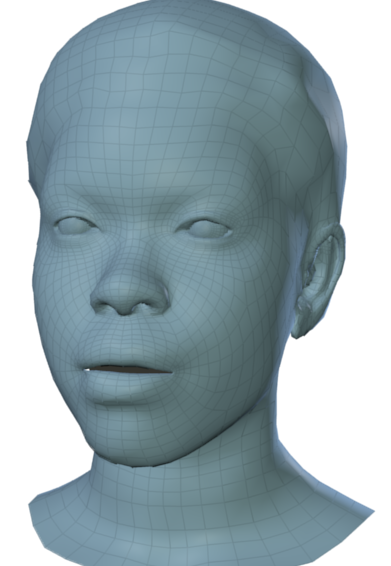} 
		\includegraphics[width=0.24\columnwidth]{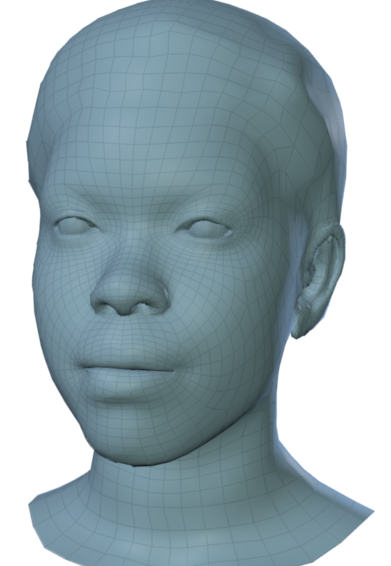} 		
	} 
	\centerline{
		\includegraphics[width=0.24\columnwidth]{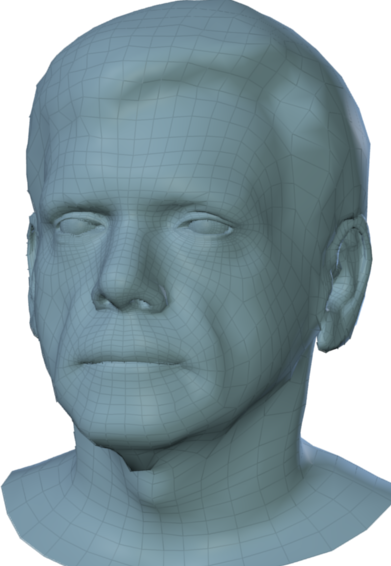} 
		\includegraphics[width=0.24\columnwidth]{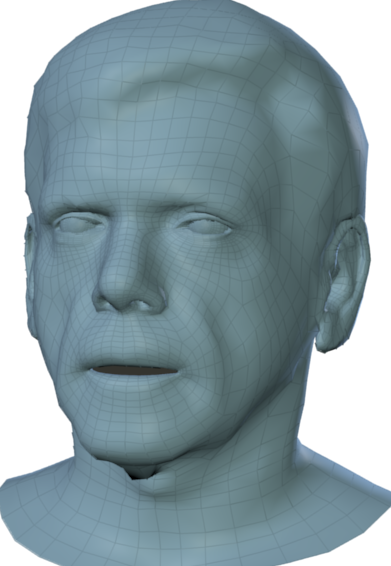} 
		\includegraphics[width=0.24\columnwidth]{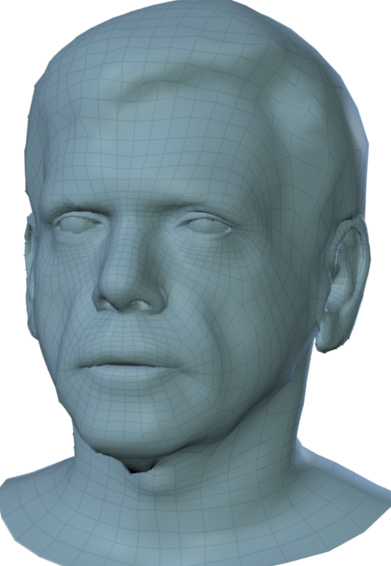} 
		\includegraphics[width=0.24\columnwidth]{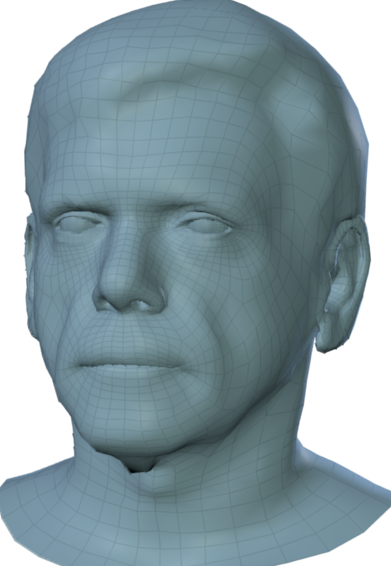} 		
	} 					
	\centerline{
		template 
		\hspace{0.26\columnwidth}	
		animation frames
		\hspace{0.2\columnwidth}	
	}
	\caption{\modelname generalizes across face shapes. Each row shows the template of a subject selected from the static BU-3DFE face database~\cite{BU-3DFE_2006} (left), and three randomly selected animation frames, driven by the same audio input (right).}
	\label{fig:generalization_across_subj}
\end{figure}

\qheading{Generalization across subjects:} Factoring identity from facial motion allows us to animate a wide range of adult faces. To show the generalization capabilities of \modelname, we select, align and pose-normalize multiple neutral scans from the BU-3DFE database~\cite{BU-3DFE_2006}, with large shape variations. Figure~\ref{fig:generalization_across_subj} shows the static template (left) and some \modelname animation frames, driven by the same audio sequence. 

\qheading{Generalization across languages:} The video shows the \modelname output for different languages. This indicates that \modelname can generalize to non-English sentences. 

\begin{figure}
	\includegraphics[width=0.95\columnwidth]{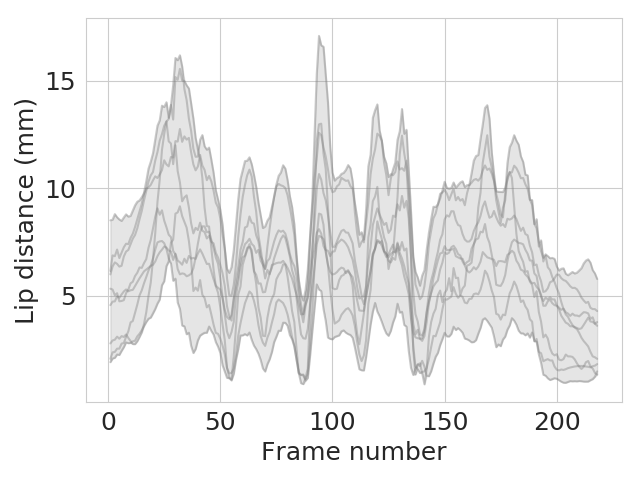}
	\caption{Distance between lower and upper lip for \modelname predictions conditioned on different subjects. The shaded region represents the space of convex combinations of the different conditions.}
	\label{fig:conditioning}
\end{figure}

\qheading{Speaker styles:} Conditioning on different subjects during inference results in different speaking styles. Stylistic differences include variation in lip articulation. Figure~\ref{fig:conditioning} shows the distance between lower and upper lip as a function of time for \modelname predictions for a random audio sequence and different conditions. This indicates that the convex combination of styles provides a wide range of different mouth amplitudes. 

We generate new intermediate speaking styles by convex combinations of conditions. Due to the linearity of the decoder, performing this convex combination in the 3D vertex space or in the 50-dimensional encoding space is equivalent. The supplementary video shows that combining styles offers animation control to synthesize a range of varying speaking styles. This is potentially useful for matching the speaking performance of a subject not seen during training. In the future, this could be estimated from video. 

\qheading{Robustness to noise:} To demonstrate robustness to noise, we combine a speech signal with different levels of noise and use the noisy signal as \modelname input. As a noise source, we use a realistic street noise sequence~\cite{soundible} added with negative gain of $36$dB (low), $24$dB (medium), $18$dB (slightly high), and $12$dB (high). Only the high noise level leads to a damped facial motion, but despite the noise, the facial animations remain plausible.

\qheading{Comparison to Karras et al.~\cite{karras2017}:} We compare \modelname to Karras et al.~\cite{karras2017}, the state-of-the-art in realistic subject-specific audio-driven facial animation. The results are shown in the supplementary video. For comparison, the authors provided us with a static mesh, to which we aligned the FLAME topology. We then use eight audio sequences from their supplementary video (including singing, spoken Chinese, an excerpt of a Barack Obama speech, and different sequences of the actor), to animate their static mesh. The supplementary video shows that, while their model produces more natural and detailed results, we can still reproduce similar facial animation without using any of their subject-specific training data. Further, Karras et al. use professional actors capable of simulating emotional speech. This enables them to add more realism in the upper face by modeling motions (i.e. eyes and eyebrows) that are more correlated with emotions than speech.

\begin{figure}[t]
	\centerline{
		\includegraphics[width=0.24\columnwidth]{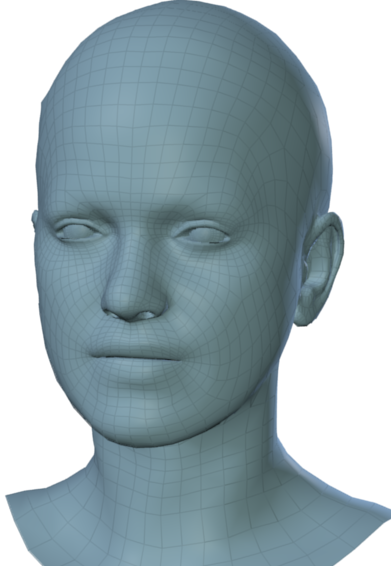} 
		\includegraphics[width=0.24\columnwidth]{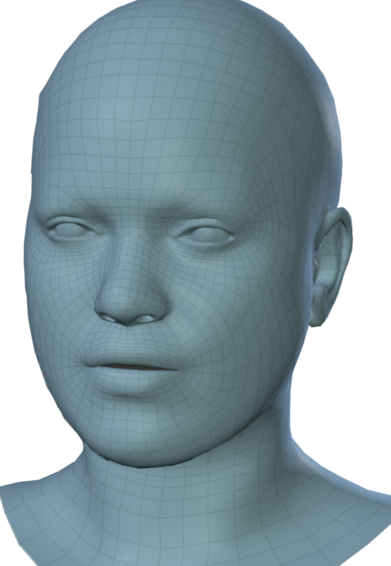} 
		\includegraphics[width=0.24\columnwidth]{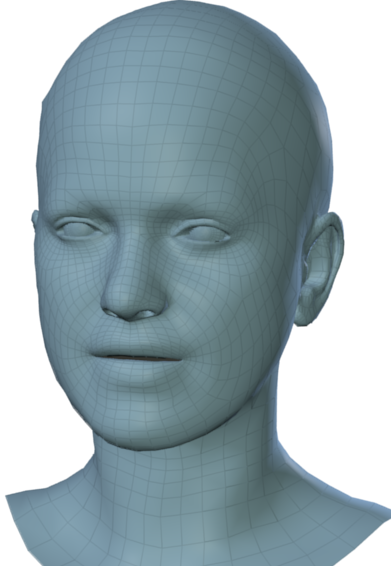} 
		\includegraphics[width=0.24\columnwidth]{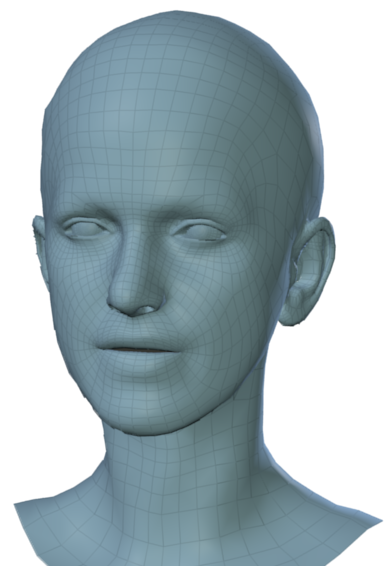} 
	} 
	\centerline{
		\includegraphics[width=0.24\columnwidth]{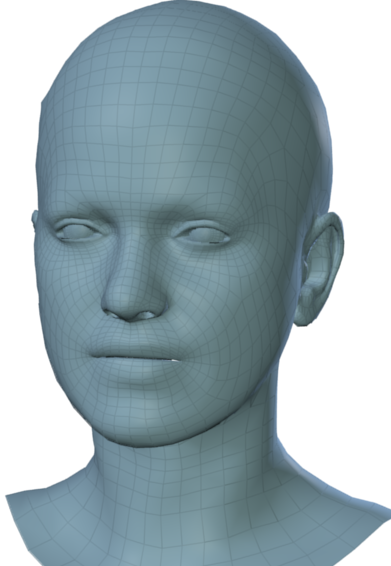} 
		\includegraphics[width=0.24\columnwidth]{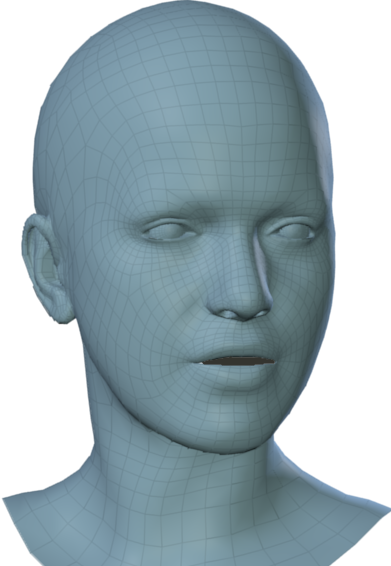} 
		\includegraphics[width=0.24\columnwidth]{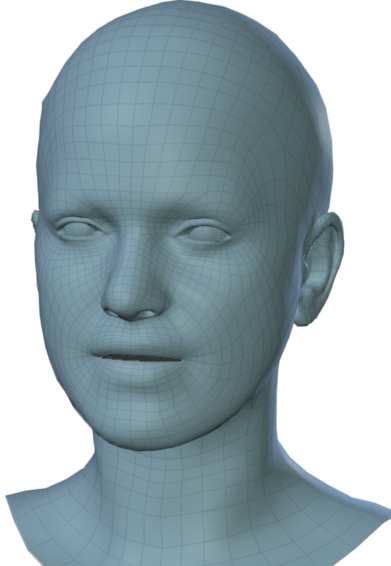} 
		\includegraphics[width=0.24\columnwidth]{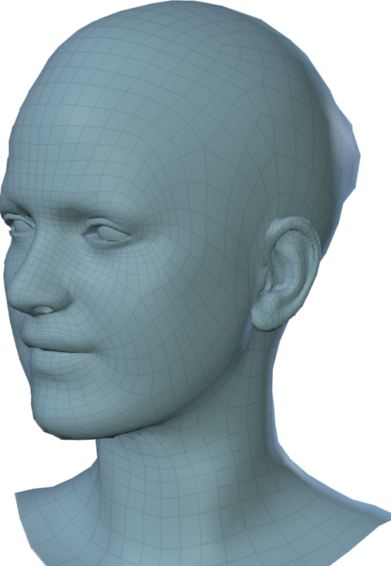} 
	} 
	\caption{Animation control. Top: varying the first identity shape components to plus two (second column) and minus two (last column) standard deviations. Bottom: varying the head pose to minus 30 degrees (second column) and plus 30 degrees (last column).}
	\label{fig:modified_shape}
\end{figure}

\qheading{Animation control:} Figure~\ref{fig:modified_shape} demonstrates the possibility of changing the identity dependent shape (top) and head pose (bottom) during animation. Both rows are driven by the same audio sequence. Despite the varying shape or pose, the facial animation looks realistic. 

\section{Discussion}

While \modelname can be used to realistically animate a wide range of adults faces from speech, it still lacks some of the details needed for conversational realism.
Upper face motions (i.e. eyes and eyebrows) are not strongly correlated with the audio~\cite{karras2017}. The causal factor is emotion, which is absent in our data due the inherent difficulty of simulating emotional speech in a controlled capture environment. Thus, \modelname learns the causal facial motions from speech, which are mostly present in the lower face. 

Non-verbal communication cues, such as head motion, are weakly correlated with the audio signal and hence are not modeled well by audio-driven techniques.
\modelname offers animators and developers the possibility to include head motion, but does not infer it from data.
A speech independent model for head motion could be used to simulate realistic results. Application specific techniques, such as dyadic interactions between animated assistants and humans require attention mechanisms that consider spatial features, such as eye tracking. 
Learning richer conversation models with expressive bodies~\cite{SMPLEx2019} is future research.

Conditioning on subject labels is one of the key aspects of \modelname that allows training across subjects. This allows a user to alter the speaking style during inference. Using data from more subjects to increase the number of different speaking styles remains a task for future work. Further experiments on mitigating or amplifying different speaking styles, or combining characteristics of different subjects also remain for future work.

\section{Conclusion}

We have presented \modelname, a simple and generic speech-driven facial
animation framework that works across a range of identities. Given an arbitrary speech
signal and a static character mesh, \modelname fully automatically outputs a realistic character animation.
\modelname leverages recent advances in speech processing and 3D face modeling in order to be subject independent. 
We train our model on a self-captured multi-subject 4D face dataset (VOCASET).
The key insights of \modelname are to factor identity from facial motion, which allows us to animate a wide range of adult faces, and to condition on subject labels, which enables us to train \modelname across multiple subjects, and to synthesize different speaker styles during test time. 
\modelname generalizes well across various speech sources, languages, and 3D face templates.
We provide optional animation control parameters to vary the speaking style and to alter the identity dependent shape and head pose during animation.
The dataset, trained model, and code are available for research purposes~\cite{VOCA_project}.

\section*{Acknowledgement}

We thank T. Alexiadis and J.~M\'{a}rquez for the data acquisition, B. Pellkofer for hardware support, A. Quiros-Ramires for support with MTurk, A. Osman for support with Tensorflow, and S. Pujades for help with finalizing the paper. We further thank Karras et al. for providing us with a static face mesh for comparison.

 MJB has received research gift funds from Intel, Nvidia, Adobe, Facebook, and Amazon. While MJB is a part-time employee of Amazon, his research was performed solely at, and funded solely by, MPI. MJB has financial interests in Amazon and Meshcapade GmbH.

\textbf{In loving memory of Daniel Cudeiro}.


{\small
\begin{thebibliography}{10}\itemsep=-1pt

\bibitem{Abadi2016}
M. Abadi, A. Agarwal, P. Barham, E. Brevdo, Z. Chen, C. Citro, G.~S. Corrado,
  A. Davis, J. Dean, M. Devin, et~al.
\newblock Tensorflow: Large-scale machine learning on heterogeneous distributed
  systems.
\newblock {\em arXiv preprint arXiv:1603.04467}, 2016.

\bibitem{Alashkar2014}
T. Alashkar, B.~Ben Amor, M. Daoudi, and S. Berretti.
\newblock A {3D} dynamic database for unconstrained face recognition.
\newblock In {\em {International Conference and Exhibition on 3D Body Scanning
  Technologies}}, 2014.

\bibitem{Alexander2009}
O. Alexander, M. Rogers, W. Lambeth, M. Chiang, and P. Debevec.
\newblock The digital emily project: Photoreal facial modeling and animation.
\newblock In {\em SIGGRAPH 2009 Courses}, pages 12:1--12:15, 2009.

\bibitem{Anderson2013}
R. Anderson, B. Stenger, V. Wan, and R. Cipolla.
\newblock Expressive visual text-to-speech using active appearance models.
\newblock In {\em Conference on Computer Vision and Pattern Recognition}, pages
  3382--3389, 2013.

\bibitem{Bolkart2015}
T. Bolkart and S. Wuhrer.
\newblock A groupwise multilinear correspondence optimization for {3D} faces.
\newblock In {\em International Conference on Computer Vision}, pages
  3604--3612, 2015.

\bibitem{Bouaziz2013}
S. Bouaziz, Y. Wang, and M. Pauly.
\newblock Online modeling for realtime facial animation.
\newblock {\em Transactions on Graphics}, 32(4):40, 2013.

\bibitem{Brand1999}
M. Brand.
\newblock Voice puppetry.
\newblock In {\em SIGGRAPH}, pages 21--28, 1999.

\bibitem{Bregler1997}
C. Bregler, M. Covell, and M. Slaney.
\newblock Video rewrite: Driving visual speech with audio.
\newblock In {\em SIGGRAPH}, pages 353--360, 1997.

\bibitem{BulatTzimiropoulos}
A. Bulat and G. Tzimiropoulos.
\newblock How far are we from solving the {2D} {\&} {3D} face alignment
  problem? (and a dataset of {230,000} {3D} facial landmarks).
\newblock In {\em International Conference on Computer Vision}, pages
  1021--1030, 2017.

\bibitem{Busso2007}
C. Busso, Z. Deng, M. Grimm, U. Neumann, and S. Narayanan.
\newblock Rigid head motion in expressive speech animation: Analysis and
  synthesis.
\newblock {\em Transactions on Audio, Speech, and Language Processing},
  15(3):1075--1086, 2007.

\bibitem{Cao2015}
C. Cao, D. Bradley, K. Zhou, and T. Beeler.
\newblock Real-time high-fidelity facial performance capture.
\newblock {\em Transactions on Graphics (Proceedings of SIGGRAPH)},
  34(4):46:1--46:9, 2015.

\bibitem{Cao2014}
C. Cao, Q. Hou, and K. Zhou.
\newblock Displaced dynamic expression regression for real-time facial tracking
  and animation.
\newblock {\em Transactions on Graphics (Proceedings of SIGGRAPH)},
  33(4):43:1--43:10, 2014.

\bibitem{Cao2014_FaceWarehouse}
C. Cao, Y. Weng, S. Zhou, Y. Tong, and K. Zhou.
\newblock Facewarehouse: A {3D} facial expression database for visual
  computing.
\newblock {\em Transactions on Visualization and Computer Graphics},
  20(3):413--425, 2014.

\bibitem{Cao2005}
Y. Cao, W.~C. Tien, P. Faloutsos, and F. Pighin.
\newblock Expressive speech-driven facial animation.
\newblock {\em Transactions on Graphics}, 24(4):1283--1302, 2005.

\bibitem{Chang2005}
Y. Chang, M. Vieira, M. Turk, and L. Velho.
\newblock Automatic {3D} facial expression analysis in videos.
\newblock In {\em Analysis and Modelling of Faces and Gestures}, pages
  293--307, 2005.

\bibitem{Chen2018}
L. Chen, Z. Li, R.~K. Maddox, Z. Duan, and C. Xu.
\newblock Lip movements generation at a glance.
\newblock In {\em European Confence on Computer Vision}, pages 538--553, 2018.

\bibitem{4DFAB2018}
S. Cheng, I. Kotsia, M. Pantic, and S. Zafeiriou.
\newblock {4DFAB}: A large scale 4d database for facial expression analysis and
  biometric applications.
\newblock In {\em CVPR}, 2018.

\bibitem{D3DFACS2011}
D. Cosker, E. Krumhuber, and A. Hilton.
\newblock A {FACS} valid {3D} dynamic action unit database with applications to
  {3D} dynamic morphable facial modeling.
\newblock In {\em International Conference on Computer Vision}, pages
  2296--2303, 2011.

\bibitem{Dale2011}
K. Dale, K. Sunkavalli, M.~K. Johnson, D. Vlasic, W. Matusik, and H. Pfister.
\newblock Video face replacement.
\newblock {\em Transactions on Graphics (Proceedings of SIGGRAPH Asia)},
  30(6):130:1--10, 2011.

\bibitem{Ding2015}
C. Ding, L. Xie, and P. Zhu.
\newblock Head motion synthesis from speech using deep neural networks.
\newblock {\em Multimedia Tools and Applications}, 74(22):9871--9888, 2015.

\bibitem{Edwards2016_JALI}
P. Edwards, C. Landreth, E. Fiume, and K. Singh.
\newblock {JALI}: An animator-centric viseme model for expressive lip
  synchronization.
\newblock {\em Transactions on Graphics (Proc. SIGGRAPH)}, 35(4):127:1--127:11,
  2016.

\bibitem{Ephrat:Siggraph:2018}
A. Ephrat, I. Mosseri, O. Lang, T. Dekel, K. Wilson, A. Hassidim, W.~T.
  Freeman, and M. Rubinstein.
\newblock Looking to listen at the cocktail party: A speaker-independent
  audio-visual model for speech separation.
\newblock {\em Transactions on Graphics}, 37(4):112:1--112:11, 2018.

\bibitem{Ezzat2002}
T. Ezzat, G. Geiger, and T. Poggio.
\newblock Trainable videorealistic speech animation.
\newblock {\em Transactions on Graphics (Proc. SIGGRAPH)}, 21(3):388--398,
  2002.

\bibitem{fan2015photo}
B. Fan, L. Wang, F.~K. Soong, and L. Xie.
\newblock Photo-real talking head with deep bidirectional {LSTM}.
\newblock In {\em International Conference on Acoustics, Speech and Signal
  Processing}, pages 4884--4888, 2015.

\bibitem{fan2016deep}
B. Fan, L. Xie, S. Yang, L. Wang, and F.~K. Soong.
\newblock A deep bidirectional {LSTM} approach for video-realistic talking
  head.
\newblock {\em Multimedia Tools and Applications}, 75(9):5287--5309, 2016.

\bibitem{B3D(AC)2010}
Gabrielle Fanelli, J{\"u}rgen Gall, Harald Romsdorfer, Thibaut Weise, and Luc
  van Gool.
\newblock A {3D} audio-visual corpus of affective communication.
\newblock {\em IEEE MultiMedia}, 12(6):591 -- 598, 2010.

\bibitem{fisher1986}
W.~M. Fisher, G.~R. Doddington, and K.~M. Goudie-Marshall.
\newblock The {DARPA} speech recognition research database: Specifications and
  status.
\newblock In {\em {DARPA} Speech Recognition Workshop}, 1986.

\bibitem{timit1993}
J.~S. Garofolo, L.~F. Lamel, W.~M. Fisher, J.~G. Fiscus, D.~S. Pallett, and
  N.~L. Dahlgren.
\newblock Darpa timit acoustic phonetic continuous speech corpus cdrom, 1993.

\bibitem{hannun2014deep}
A. Hannun, C. Case, J. Casper, B. Catanzaro, G. Diamos, E. Elsen, R. Prenger,
  S. Satheesh, S. Sengupta, A. Coates, et~al.
\newblock Deep speech: Scaling up end-to-end speech recognition.
\newblock {\em arXiv preprint arXiv:1412.5567}, 2014.

\bibitem{Hermansky1994}
H. Hermansky and N. Morgan.
\newblock {RASTA} processing of speech.
\newblock {\em Transactions on Speech and Audio Processing}, 2(4):578--589,
  1994.

\bibitem{Hong2002}
P. Hong, Z. Wen, and T.~S. Huang.
\newblock Real-time speech-driven face animation with expressions using neural
  networks.
\newblock {\em Transactions on Neural Networks}, 13(4):916--927, 2002.

\bibitem{kakumanu2001speech}
P. Kakumanu, R. Gutierrez-Osuna, A. Esposito, R. Bryll, A. Goshtasby, and O.
  Garcia.
\newblock Speech driven facial animation.
\newblock In {\em Proceedings of the 2001 workshop on Perceptive user
  interfaces}, pages 1--5, 2001.

\bibitem{karras2017}
T. Karras, T. Aila, S. Laine, A. Herva, and J. Lehtinen.
\newblock Audio-driven facial animation by joint end-to-end learning of pose
  and emotion.
\newblock {\em Transactions on Graphics (Proc. SIGGRAPH)}, 36(4):94, 2017.

\bibitem{kingma2014adam}
D. Kingma and J. Ba.
\newblock Adam: A method for stochastic optimization.
\newblock {\em arXiv preprint arXiv:1412.6980}, 2014.

\bibitem{Laine2017}
S. Laine, T. Karras, T. Aila, A. Herva, S. Saito, R. Yu, H. Li, and J.
  Lehtinen.
\newblock Production-level facial performance capture using deep convolutional
  neural networks.
\newblock In {\em SIGGRAPH / Eurographics Symposium on Computer Animation},
  pages 10:1--10:10, 2017.

\bibitem{Li2010}
H. Li, T. Weise, and M. Pauly.
\newblock Example-based facial rigging.
\newblock 29(4):32, 2010.

\bibitem{flame2017}
T. Li, T. Bolkart, M.~J. Black, H. Li, and J. Romero.
\newblock Learning a model of facial shape and expression from {4D} scans.
\newblock {\em ACM Transactions on Graphics}, 36(6), 2017.

\bibitem{Liu2015}
Y. Liu, F. Xu, J. Chai, X. Tong, L. Wang, and Q. Huo.
\newblock Video-audio driven real-time facial animation.
\newblock {\em Transactions on Graphics (Proc. SIGGRAPH Asia)},
  34(6):182:1--182:10, 2015.

\bibitem{Mori1970}
M. Mori.
\newblock {Bukimi no tani [the uncanny valley]}.
\newblock {\em Energy}, 7:33--35, 1970.

\bibitem{mozillaDeepSpeech}
Mozilla.
\newblock {Project DeepSpeech}.
\newblock \url{https://github.com/mozilla/DeepSpeech}, 2017.

\bibitem{SMPLEx2019}
G. Pavlakos, V. Choutas, N. Ghorbani, T. Bolkart, A.~A.~A. Osman, D. Tzionas,
  and M.~J. Black.
\newblock Expressive body capture: 3d hands, face, and body from a single
  image.
\newblock In {\em Computer Vision and Pattern Recognition}, 2019.

\bibitem{Pham2017}
H. Pham and V. Pavlovic.
\newblock Speech-driven {3D} facial animation with implicit emotional
  awareness: A deep learning approach.
\newblock In {\em Conference on Computer Vision and Pattern Recognition
  Workshop}, 2017.

\bibitem{squad2016}
P. Rajpurkar, J. Zhang, K. Lopyrev, and P. Liang.
\newblock Squad: {100,000+} questions for machine comprehension of text.
\newblock 2016.

\bibitem{CoMA2018}
A. Ranjan, T. Bolkart, S. Sanyal, and M.~J. Black.
\newblock Generating {3D} faces using convolutional mesh autoencoders.
\newblock In {\em European Confence on Computer Vision}, pages 725--741, 2018.

\bibitem{Sako2000}
S. Sako, K. Tokuda, T. Masuko, T. Kobayashi, and T. Kitamura.
\newblock {HMM}-based text-to-audio-visual speech synthesis.
\newblock In {\em International Conference on Spoken Language Processing},
  pages 25--28, 2000.

\bibitem{Salvi2009}
G. Salvi, J. Beskow, S. Al~Moubayed, and B. Granstr{\"o}m.
\newblock Synface---speech-driven facial animation for virtual speech-reading
  support.
\newblock {\em EURASIP Journal on Audio, Speech, and Music Processing},
  2009(1), 2009.

\bibitem{Bosphorus2008}
A. Savran, N. Alyu\"{o}z, H. Dibeklioglu, O. Celiktutan, B. G\"{o}kberk, B.
  Sankur, and L. Akarun.
\newblock Bosphorus database for {3D} face analysis.
\newblock In {\em Biometrics and Identity Management}, pages 47--56, 2008.

\bibitem{shimba2015talking}
T. Shimba, R. Sakurai, H. Yamazoe, and J.-H. Lee.
\newblock Talking heads synthesis from audio with deep neural networks.
\newblock In {\em System Integration}, pages 100--105, 2015.

\bibitem{soundible}
soundible.com.
\newblock \url{http://soundbible.com/tags-city.html}.

\bibitem{SumnerPopovic2004}
R.W. Sumner and J. Popovi\'{c}.
\newblock Deformation transfer for triangle meshes.
\newblock {\em Transactions on Graphics (Proceedings of SIGGRAPH)},
  23(3):399--405, 2004.

\bibitem{obama2017}
S. Suwajanakorn, S.~M. Seitz, and I. Kemelmacher-Shlizerman.
\newblock Synthesizing {Obama}: learning lip sync from audio.
\newblock {\em Transactions on Graphics (Proc. SIGGRAPH)}, 36(4):95, 2017.

\bibitem{taylor2016audio}
S. Taylor, A. Kato, B. Milner, and I. Matthews.
\newblock Audio-to-visual speech conversion using deep neural networks.
\newblock 2016.

\bibitem{taylor2017deep}
S. Taylor, T. Kim, Y. Yue, M. Mahler, J. Krahe, A.~G. Rodriguez, J. Hodgins,
  and I. Matthews.
\newblock A deep learning approach for generalized speech animation.
\newblock {\em Transactions on Graphics}, 36(4):93, 2017.

\bibitem{taylor2012dynamic}
S.~L. Taylor, M. Mahler, B.-J. Theobald, and I. Matthews.
\newblock Dynamic units of visual speech.
\newblock In {\em SIGGRAPH/Eurographics conference on Computer Animation},
  pages 275--284, 2012.

\bibitem{Thies2016_Face2Face}
J. Thies, M. Zollh{\"o}fer, M. Stamminger, C. Theobalt, and M. Nie{\ss}ner.
\newblock {Face2Face: Real-time Face Capture and Reenactment of RGB Videos}.
\newblock In {\em Computer Vision and Pattern Recognition}, 2016.

\bibitem{WaveNet2016}
A. van~den Oord, S. Dieleman, H. Zen, . Simonyan, O. Vinyals, A. Graves, N.
  Kalchbrenner, A. Senior, and . Kavukcuoglu.
\newblock Wavenet: A generative model for raw audio.
\newblock {\em CoRR}, abs/1609.03499, 2016.

\bibitem{Vlasic2005}
D. Vlasic, M. Brand, H. Pfister, and J. Popovi\'{c}.
\newblock Face transfer with multilinear models.
\newblock {\em Transactions on Graphics (Proceedings of SIGGRAPH)},
  24(3):426--433, 2005.

\bibitem{VOCA_project}
VOCA.
\newblock \url{http://voca.is.tue.mpg.de}, 2019.

\bibitem{Wang2011}
L. Wang, W. Han, F. Soong, and Q. Huo.
\newblock Text driven 3d photo-realistic talking head.
\newblock In {\em Conference of the International Speech Communication
  Association, INTERSPEECH}, pages 3307--3308, 2011.

\bibitem{Weise2011}
T. Weise, S. Bouaziz, H. Li, and M. Pauly.
\newblock Realtime performance-based facial animation.
\newblock {\em Transactions on Graphics (Proceedings of SIGGRAPH)},
  30(4):77:1--77:10, 2011.

\bibitem{Wu2016}
C. Wu, D. Bradley, M. Gross, and T. Beeler.
\newblock An anatomically-constrained local deformation model for monocular
  face capture.
\newblock {\em Transactions on Graphics (Proc. SIGGRAPH)}, 2016.

\bibitem{XieLiu2007}
L. Xie and Z.-Q. Liu.
\newblock Realistic mouth-synching for speech-driven talking face using
  articulatory modelling.
\newblock {\em Transactions on Multimedia}, 9(3):500--510, 2007.

\bibitem{Yang2012}
F. Yang, L. Bourdev, J. Wang, E. Shechtman, and D. Metaxas.
\newblock Facial expression editing in video using a temporally-smooth
  factorization.
\newblock In {\em Conference on Computer Vision and Pattern Recognition}, pages
  861--868, 2012.

\bibitem{BU-4DFE2008}
L. Yin, X. Chen, Y. Sun, T. Worm, and M. Reale.
\newblock A high-resolution {3D} dynamic facial expression database.
\newblock In {\em International Conference on Automatic Face and Gesture
  Recognition}, pages 1--6, 2008.

\bibitem{BU-3DFE2006}
L. Yin, X. Wei, Y. Sun, J. Wang, and M.~J. Rosato.
\newblock A {3D} facial expression database for facial behavior research.
\newblock In {\em International Conference on Automatic Face and Gesture
  Recognition}, pages 211--216, 2006.

\bibitem{BU-3DFE_2006}
L. Yin, X. Wei, Y. Sun, J. Wang, and M.~J. Rosato.
\newblock A {3D} facial expression database for facial behavior research.
\newblock In {\em International Conference on Automatic Face and Gesture
  Recognition}, pages 211--216, 2006.

\bibitem{Zhang2013}
X. Zhang, L. Wang, G. Li, F. Seide, and F.~K. Soong.
\newblock A new language independent, photo-realistic talking head driven by
  voice only.
\newblock In {\em {INTERSPEECH}}, pages 2743--2747, 2013.

\bibitem{BP4D-Spontaneous2014}
X. Zhang, L. Yin, J.~F. Cohn, S. Canavan, M. Reale, A. Horowitz, P. Liu, and
  J.~M. Girard.
\newblock {BP4D}-spontaneous: a high-resolution spontaneous {3D} dynamic facial
  expression database.
\newblock {\em Image and Vision Computing}, 32(10):692 -- 706, 2014.

\bibitem{MMSE2016}
Z. Zhang, J.~M. Girard, Y. Wu, X. Zhang, P. Liu, U. Ciftci, S. Canavan, M.
  Reale, A. Horowitz, H. Yang, J.~F. Cohn, Q. Ji, and L. Yin.
\newblock Multimodal spontaneous emotion corpus for human behavior analysis.
\newblock In {\em CVPR}, pages 3438--3446, 2016.

\bibitem{Zhou2018}
Y. Zhou, Y. Xu, C. Landreth, E. Kalogerakis, S. Maji, and K. Singh.
\newblock Visemenet: Audio-driven animator-centric speech animation.
\newblock {\em Transactions on Graphics}, 37(4):161:1--161:10, 2018.

\end{thebibliography}

}
\end{document}